\documentclass{bvm} 

\begin{document}

\newcommand{\bvmyear}{2026}

\selectlanguage{english} 

\title{MedDIFT: Multi-Scale Diffusion-Based Correspondence
in 3D Medical Imaging}


\titlerunning{MedDIFT: Diffusion-Based Correspondence}

\author{
	\fname{Xingyu} \lname{Zhang} \inst{1} \affiliation{TUM} \city{Garching} \country{Germany} \authorsEmail{xingyu.zhang@tum.de} \street{Lichtenbergstr.} \housenumber{2a} \zipcode{85748},
  \fname{Anna} \lname{Reithmeir} \inst{1,2,3} \affiliation{TUM} \city{Garching} \country{Germany} \authorsEmail{anna.reithmeir@tum.de} \street{Lichtenbergstr.} \housenumber{2a} \zipcode{85748},
  \fname{Fryderyk} \lname{Kögl} \inst{1,2,3,4} \affiliation{TUM} \city{Garching} \country{Germany} \authorsEmail{fryderyk.koegl@tum.de} \street{Lichtenbergstr.} \housenumber{2a} \zipcode{85748},
  \fname{Rickmer} \lname{Braren} \inst{4} \affiliation{Klinikum Rechts der Isar}\city{Munich} \country{Germany} \authorsEmail{rbraren@tum.de} \street{Ismaninger Str.} \housenumber{22} \zipcode{81675},
  \fname{Julia~A.} \lname{Schnabel} \inst{1,2,3,5} \affiliation{TUM} \city{Garching} \country{Germany} \authorsEmail{julia.schnabel@tum.de} \street{Lichtenbergstr.} \housenumber{2a} \zipcode{85748},
  \fname{Daniel~M.} \lname{Lang} \inst{1,2} \affiliation{Helmholtz} \city{Neuherberg} \country{Germany} \authorsEmail{lang@helmholtz-munich.de} \street{Ingolstädter Landstr.} \housenumber{1} \zipcode{85764} \isResponsibleAuthor
}
\authorrunning{Zhang et al.}
\institute{
\inst{1} School of Computation, Information and Technology, Technical University of Munich\\
\inst{2} Institute of Machine Learning in Biomedical Imaging, Helmholtz Munich\\
\inst{3} Munich Center for Machine Learning (MCML)\\
\inst{4} Institute for Diagnostic and Interventional Radiology, Klinkum Rechts der Isar\\
\inst{5} School of Biomedical Engineering \& Imaging Sciences, King’s College London
}
\email{anna.reithmeir@tum.de}

\maketitle

\begin{abstract}
Accurate spatial correspondence between medical images is essential for longitudinal analysis, lesion tracking, and image-guided interventions. 
Medical image registration methods rely on local intensity-based similarity measures, which fail to capture global semantic structure and often yield mismatches in low-contrast or anatomically variable regions.
Recent advances in diffusion models suggest that their intermediate representations encode rich geometric and semantic information. 
We present MedDIFT, a training-free 3D correspondence framework that leverages multi-scale features from a pretrained latent medical diffusion model as voxel descriptors.
MedDIFT fuses diffusion activations into rich voxel-wise descriptors and matches them via cosine similarity, with an optional local-search prior. 
On a publicly available lung CT dataset, MedDIFT shows promising capability in identifying anatomical correspondence without requiring any task-specific model training.
Ablation experiments confirm that multi-level feature fusion and modest diffusion noise improve performance.
Code is available online\footnote{https://github.com/merlinxyz/MedDIFT}.
\end{abstract}

\section{Introduction}
In medical imaging, accurate correspondences are fundamental for various applications including longitudinal disease assessment, lesion tracking, and therapy planning. 
They are typically obtained through image registration, which estimates dense mappings between volumes acquired at different time points, across patients, or between modalities.

Classical and learning-based registration algorithms maximize local intensity-based similarity measures such as cross-correlation or mutual information. 
While effective in regions with clear intensity structure, these measures operate purely on local appearance and can fail in areas with low contrast, artifacts, or anatomical variability.

Recent advances in diffusion models \cite{3832-ho2020denoising} have revealed that their intermediate feature activations encode rich, global semantic structures. 
The diffusion features framework (DIFT) \cite{3832-tang2023emergent} demonstrated that such diffusion-derived representations enable robust semantic, geometric, and temporal matching across (2D) natural images without task-specific training.
More recently, diffusion features have been explored in the context of image registration \cite{3832-tursynbek2025guiding}. However, the method relies on a 2D natural image pretrained diffusion model, which was not specifically designed for medical images, and is not inherently three-dimensional.

Motivated by this research gap, we advance DIFT to the medical domain and propose MedDIFT, the first framework to leverage diffusion features from a pre-trained 3D medical diffusion model for establishing voxel correspondences in 3D medical images.
MedDIFT extracts multi-scale semantic descriptors from the pre-trained latent diffusion model MAISI ~\cite{3832-guo2025maisi} and matches corresponding voxels using cosine similarity \emph{without any weight optimization or task-specific training}.
By capturing both local and global anatomical semantics, MedDIFT bridges the gap between conventional local-intensity registration and rich learned features.
Our key contributions are:
\begin{itemize}
\item We introduce MedDIFT, a training-free framework for 3D medical voxel correspondence based on medical diffusion features;
\item We show MedDIFT's competitive matching accuracy relative to a learning-based registration approach, and we identify opportunities for optimization.
\end{itemize}

\begin{figure}
\includegraphics[width=\textwidth]{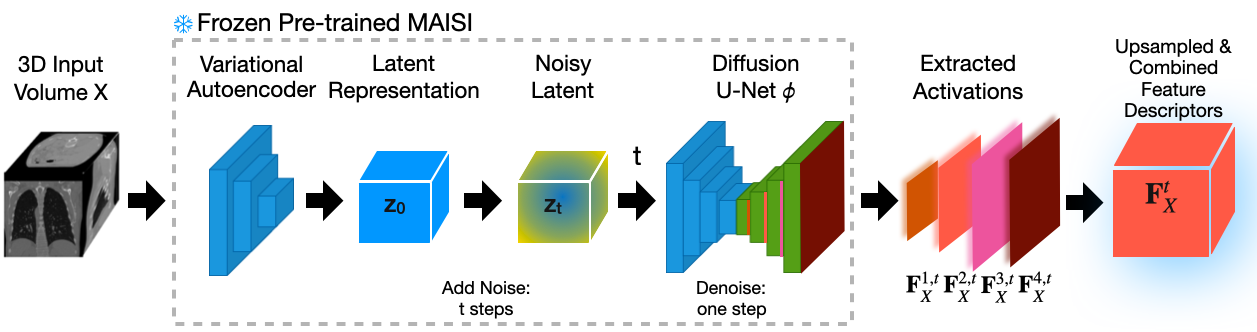}
\caption{MedDIFT derives diffusion features from the pretrained MAISI \cite{3832-guo2025maisi} model and fuses them into semantic feature descriptors.} \label{3832-fig:Schema}
\altText{MedDIFT derives diffusion features from the pretrained MAISI \cite{3832-guo2025maisi} model and fuses them into semantic feature descriptors. Description see text.}
\end{figure}


\section{Materials and methods}

\subsection{Method}

Given a pair of 3D medical images $\mathbf{A},\mathbf{B}\in\mathbb{R}^{H\times W\times D}$, our goal is to establish voxel correspondences $y\in\mathbf{B}$ for voxel $x\in\mathbf{A}$ by matching semantically similar anatomical regions.
Unlike image registration methods that rely on local intensity similarity or handcrafted descriptors, MedDIFT constructs diffusion-based voxel descriptors that capture global semantic context derived from a pretrained medical diffusion model. 
The framework operates without any finetuning or task-specific training and consists of three stages: 
(i) Multi-scale diffusion feature extraction, (ii) feature fusion into voxel descriptors, and (iii) correspondence matching. Each stage is described below. 

\paragraph{Diffusion feature extraction}
We build upon the MAISI latent diffusion model \cite{3832-guo2025maisi} that was trained to generate 3D CT images.
In latent diffusion models~\cite{3832-rombach2022high}, a clean latent $z_o$ is progressively noised to $z_t$ following a predefined noise schedule  $\{\alpha_t\}_{t=1}^T$:
\begin{equation}
z_t = \sqrt{\alpha_t}z_0 + \sqrt{1-\alpha_t}\epsilon, \quad \epsilon \sim \mathcal{N}(0, I).
\end{equation}
During the denoising process, intermediate activations at different network layers and timesteps $t$ capture progressively refined semantic information.

We follow the DIFT approach to extract diffusion features for an input image $\mathbf{X}$ (see Fig.~\ref{3832-fig:Schema}). 
First, we encode $\mathbf{X}$ into a latent representation $\mathbf{z}_0$ using MAISI's variational autoencoder. 
Gaussian noise is added to simulate the forward diffusion process, from which we obtain the noisy latent $\mathbf{z}_t$ at timestep $t$. 
This noisy latent is passed through the frozen diffusion U-Net $\phi$, which performs one denoising step, similar to \cite{3832-tang2023emergent}. 
From each decoder block $l$, we extract intermediate activations $\mathbf{F}^{l,t}_X = \phi^l(z_t) \in\mathbb{R}^{C_l\times H_l\times W_l\times D_l}$, where $t$ is the noising step that controls semantic abstraction. 
Features are extracted independently for both input images $\mathbf{A}$ and $\mathbf{B}$.

\paragraph{Multi-scale descriptor construction}
\label{3832-sec:constr}
The extracted features $\mathbf{F}^{l,t}_X$ vary in spatial resolution for different $l$: 
$\mathbf{F}^{1,t}_X,\mathbf{F}^{2,t}_X,\mathbf{F}^{3,t}_X,\mathbf{F}^{4,t}_X$ correspond to feature tensors of size $\frac{1}{16}, \frac{1}{8}, \frac{1}{4}, \frac{1}{4}$ of the input volume, respectively.
To obtain unified diffusion descriptors $\mathbf{F}^t_A,\mathbf{F}^t_B$ for the images $\mathbf{A}, \mathbf{B}$, all feature maps are tri-linearly upsampled to the original image resolution, $L2$-normalized, and finally concatenated across levels.

\paragraph{Correspondence matching}

Given a query voxel coordinate $p\in\Omega_A$, its corresponding voxel coordinate $q^*\in\Omega_B$ is determined by maximizing the cosine similarity between the diffusion descriptors:
\begin{equation}
    q^* = \arg\max_{q \in \Omega_B} \frac{\mathbf{F}_A(p) \cdot \mathbf{F}_B(q)}{\|\mathbf{F}_A(p)\|_2 \, \|\mathbf{F}_B(q)\|_2},
\end{equation}
where $\Omega_B$ denotes the voxel coordinate set of $\mathbf{B}$. 
Optionally, the search space can be restricted to a local neighborhood $\Omega_B'\subset\Omega_B$ around the mapped coordinate of $p$ to reduce computation and exclude implausible matches. 
This is particularly useful when the images are already rigidly pre-aligned and $q^*$ is assumed to lie within a small neighborhood of $p$.

\subsection{Experimental setup}

We first conduct ablation studies on the training dataset to identify the optimal configuration of MedDIFT, and then compare our correspondence matching approach against correspondences estimated with two state-of-the-art registration methods on the validation dataset.

\emph{Ablation studies:}
We analyze the influence of (i) the number of noising timesteps $t$ and (ii) the feature levels used for similarity computation.
While DIFT \cite{3832-tang2023emergent} only investigates features extracted from a single network layer, we additionally explore multi-level feature fusion (see Sec.~\ref{3832-sec:constr}), combining feature activations extracted from multiple decoder layers.

\emph{Comparison to state-of-the-art methods:}
We compare our approach to two registration methods: The NiftyReg implementation \cite{3832-modat2010fast} of the conventional B-spline free-form deformations (FFD) registration \cite{3832-rueckert2002nonrigid} and UniGradICON \cite{3832-tian2024unigradicon}, a recent deep learning foundation model for medical image registration.
Since both methods estimate deformation fields rather than explicit correspondences, keypoint correspondences are obtained by warping source keypoints with the predicted deformation. NiftyReg is run with default parameters using normalized mutual information as the similarity metric, while UniGradICON is evaluated with its default settings.

Performance is measured as the Euclidean distance between predicted and ground-truth keypoints in physical space (millimeters).
Since image pairs can contain varying numbers of keypoints, we report two metrics: the \emph{case mean error} is calculated as the average of per-case mean errors, and the \emph{keypoint mean error}, computed as the average over all keypoints across all cases.

\begin{figure}
\includegraphics[width=\textwidth]{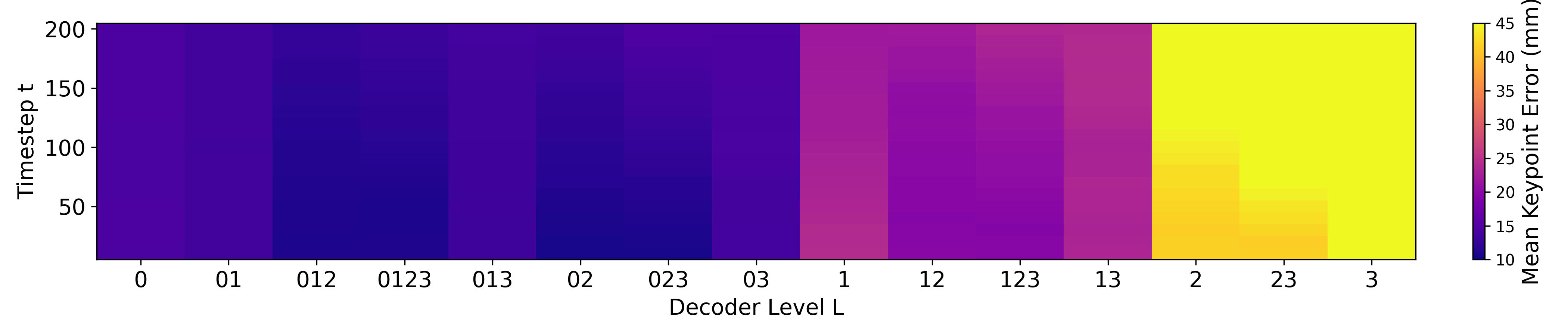}
\caption{Heatmap of mean keypoint error (in mm) across different decoder levels ($l$) and diffusion timesteps ($t$).
  Fusion of feature maps from multiple levels are indicated as, for instance, $012$ where levels $0$, $1$, and $2$ are combined. 
}
\altText{Heatmap of mean keypoint error (in mm) across different decoder levels ($l$) and diffusion timesteps ($t$).
  Fusion of feature maps from multiple levels are indicated as, for instance, $012$ where levels $0$, $1$, and $2$ are combined. 
  Errors increase for for large timesteps, while performance improves with fusing multi-level features.
}
\label{3832-fig:t_level}
\end{figure}

\subsection{Data and implementation}
Experiments are performed on the Learn2Reg Lung CT dataset~\cite{3832-hering2022learn2reg}, which comprises paired intra-patient inspiratory and expiratory chest CT scans with annotated keypoints. For each keypoint in the inspiratory scan, the correspondence in the expiratory scan is predicted.
Volumes were intensity-normalized to $[0,1]$ and resized to multiples of 128, as required by MAISI.
For our ablation study, 100 random keypoints per training case are sampled to reduce computational cost. 
For the final evaluation, all validation cases and keypoints are used.
For the restricted search variant within a neighborhood (MedDIFT-Box), we limit the search space to a bounding box defined by the 95th percentile of the initial training keypoint distances.

\section{Results}

\emph{Ablation study:}
Figure~\ref{3832-fig:t_level} shows the heatmap of mean keypoint error across varying diffusion timesteps $t$ and feature levels $l$.
Performance improves with fusing multi-level features over using only single-layer features.
Higher errors are obtained when level $0$ is not included.
Errors furthermore increase for large timesteps where strong noise is added to the latent representations.
Based on these observations, we select features from all four decoder levels and $t=20$ for subsequent experiments.

\begin{table}
\centering
\caption{Quantitative results on the Learn2Reg Lung CT test set \cite{3832-hering2022learn2reg}: 
Mean errors are reported across cases and keypoints. 
Lower values indicate better correspondence accuracy.}
\label{3832-tab:results}
\begin{tabular*}{\textwidth}{@{\extracolsep\fill}lll}
\hline
\textbf{Method} & \textbf{Case Mean $\pm$ Std (mm)} & \textbf{Keypoint Mean $\pm$ Std (mm)} \\
\hline
Nifty-Reg & $\mathbf{5.98 \pm 3.25}$ & $\mathbf{5.99 \pm 7.71}$ \\
uniGradICON & $10.03 \pm 2.60$ & $9.84 \pm 15.12$ \\
\hline
MedDIFT (ours) & $10.47 \pm 3.54$ & $10.79 \pm 12.40$ \\
MedDIFT-Box (ours) & $9.97 \pm 3.38$ & $10.21 \pm 9.56$ \\
\hline
\end{tabular*}
\end{table}

\begin{figure}[ht] 
\centering
\caption{Qualitative example. The projection onto the coronal slice of an example source keypoint (green) and
the estimated (red) and ground truth correspondences (blue) are shown, together with the similarity score map for the respective slice.
}
\altText{Qualitative example. The projection onto the coronal slice of an example source keypoint (left lung) and
the estimated and ground truth correspondences are shown, together with the similarity score map for the respective slice.
}
\includegraphics[width=\textwidth]{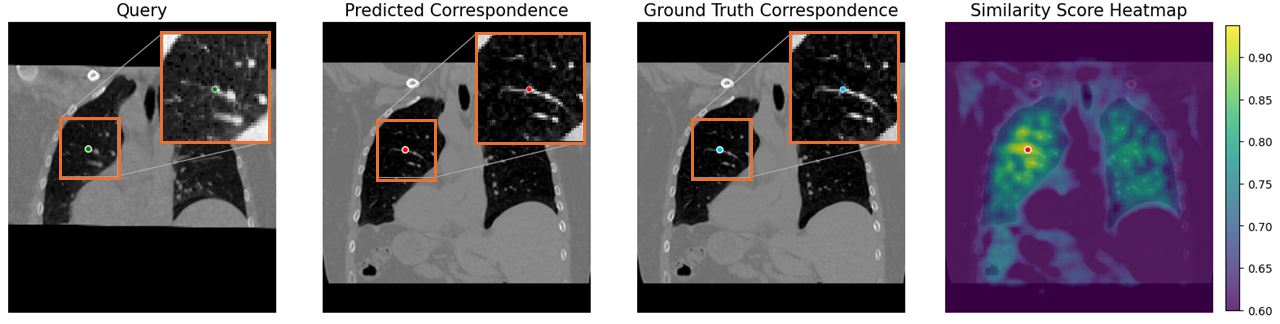}
\label{3832-fig:qualitative}
\end{figure}

\emph{Method comparison:}
Tab.~\ref{3832-tab:results} summarizes the quantitative results on the Learn2Reg Lung CT validation set. NiftyReg achieves the lowest mean error among all evaluated approaches. Compared with uniGradICON, MedDIFT achieves comparable performance.
While a higher keypoint mean error is observed, it exhibits a lower keypoint-wise standard deviation, indicating greater stability. 
When incorporating the restricted search space with a bounding box (MedDIFT-Box), the mean case error decreases further. 
A qualitative example is visualized in Fig.~\ref{3832-fig:qualitative}, with one corresponding coordinate pair projected onto the coronal slice of the query voxel. Moreover, the cosine similarity map, representing the feature similarity between the target voxel coordinates and the specified query voxel coordinate, is visualized as an overlay.

\section{Discussion}

We introduce MedDIFT, the first method to establish spatial correspondence between medical images by leveraging diffusion-based features extracted from a model pretrained on volumetric medical data, operating fully in three dimensions. Our experiments demonstrate that semantic representations extracted in this manner may provide a promising alternative to traditional intensity-based similarity measures in image registration. Although MedDIFT does not consistently surpass existing methods, it achieves competitive correspondence accuracy while operating without any task-specific training. 
A key observation is  that for medical imaging, multi-scale feature fusion consistently improves results, likely due to the combined coarse semantic and fine spatial information. Moreover, we find that moderate diffusion noise ($t\in[10,40]$) produces the best results, consistent with prior findings in the natural imaging domain~\cite{3832-tang2023emergent,3832-tursynbek2025guiding}. 

Future work will explore fine-tuning feature extractors, enhancing multi-scale feature fusion strategies, and integrating MedDIFT into registration or multimodal correspondence frameworks.

\begin{acknowledgement}
A.R. and J.A.S. acknowledge funding from the Free State of Bavaria (StMGP) as part of the project "KI-gestützte multi-modale Diagnostik und stratifizierte Therapie für Endometriose (EndoKI)".
F.K. acknowledges funding from the Federal Ministry of Education and Research (BMBF, Grant Nr. 01ZZ2315B and 01KX2021) and the Bavarian Cancer Research Center (BZKF, Lighthouse AI and Bioinformatics).
D.M.L. and J.A.S. received funding from HELMHOLTZ IMAGING, a platform of the Helmholtz Information \& Data Science Incubator. 
\end{acknowledgement}

\printbibliography

\end{document}